# Image and Depth from a Single Defocused Image Using Coded Aperture Photography

Mina Masoudifar [a], Hamid Reza Pourreza [a]
[a] Department of Computer Engineering, Ferdowsi University of Mashhad, Mashhad, Iran

## Abstract

*Depth from defocus and defocus deblurring from a single image are two challenging problems that are derived from the finite depth of field in conventional cameras. Coded aperture imaging is one of the techniques that is used for improving the results of these two problems. Up to now, different methods have been proposed for improving the results of either defocus deblurring or depth estimation. In this paper, a multi-objective function is proposed for evaluating and designing aperture patterns with the aim of improving the results of both depth from defocus and defocus deblurring. Pattern evaluation is performed by considering the scene illumination condition and camera system specification. Based on the proposed criteria, a single asymmetric pattern is designed that is used for restoring a sharp image and a depth map from a single input. Since the designed pattern is asymmetric, defocus objects on the two sides of the focal plane can be distinguished. Depth estimation is performed by using a new algorithm, which is based on image quality assessment criteria and can distinguish between blurred objects lying in front or behind the focal plane. Extensive simulations as well as experiments on a variety of real scenes are conducted to compare our aperture with previously proposed ones.*

## 1. Introduction

When a scene is imaged by a limited depth of field camera, objects at different depths of a scene are registered with various amount of defocus blur. Depth from defocus (DFD) is a method that recovers the depth information by estimating the amount of blur in different areas of a captured image. First ideas of DFD were introduced in[1],[2]. Afterward, various techniques have been proposed that use single image [3-6] or multiple images [7-10].

Single image DFDs usually estimate blur scale by supposing some prior information about PSF (Point Spread Function)[3], texture[5] or color information[6]. Multiple image DFDs are more variant and use different techniques to extract depth information. Some methods capture two or more images from a single viewpoint with different focus settings or different sizes of aperture[1, 2, 9, 11].Other methods use two or more images from different viewpoints such as stereo vision with the same focus setting[12] or different focal settings[13].

Despite achieving good results in DFD techniques with conventional apertures, there are some drawbacks because of the inherent limitation of circular apertures. For example, single image DFD methods and even some of the multiple image DFD methods cannot distinguish between defocused objects placed in the before and after focal plane. In addition, in single image DFD methods, lower depth of field, which provides better depth discrimination ability, is obtained with a cost of losing image quality. In the larger size of blur, most of image frequencies are lost. Therefore, estimating depth as well as deblurring are more ambiguous and vulnerable to image-noise [14].

Coded aperture photography is a method for modifying defocus pattern produced by lens. By using a coded mask on lens, the shape of PSF is changed. Up to now, different patterns of masks have been proposed for improving the results of depth estimation[14-16], defocus deblurring[17-19] or both of them[13, 20].
Hiura et al.[21] use multiple images that are taken by a single aperture pattern from a single viewpoint yet different focus setting. Zhou et al.[20] design a pair of aperture masks. Two images are taken from a single viewpoint and similar focus setting with different asymmetric aperture patterns. Defocus points lying in front or behind the focal plane are distinguishable. According to spectral properties of the two proposed masks, an all-focus image can also be restored. In real applications, a programmable aperture is needed to guarantee not to change the viewpoint of two captured images. Otherwise, images should be first registered, and then depth estimation algorithm is applied. Takeda et al.[13] use stereo imaging with single aperture pattern yet different focal setting to improve the results of depth estimation in [20].

Levin et al. [15] design a single symmetric pattern with the aim of increasing depth discrimination ability. Kullback-Leibler divergence between different sizes of blur is used to rank aperture patterns. A full-search of all binary masks is used for finding the best symmetric pattern. An efficient deblurring algorithm is also used to create high quality deblurred results. Since the proposed mask is symmetric, before and after focal plane cannot be differentiated.

Sellent et al.[16] define a function in the spatial domain for aperture pattern evaluation. A parametric maximization problem is defined to find a pattern that makes the most possible difference among the images which are blurred with different sizes of blur. Solving the problem results in non-binary patterns that can be pruned to binary forms. This approach is also exploited to find asymmetric patterns that are suitable to discriminate front and back of the focal plane[14].

In this paper, we search for a pattern to be appropriate for both depth estimation and deblurring. For evaluating a pattern, expected deblurring error is computed in two different states: deblurring with correct scale PSF and deblurring with incorrect scales. Our goal is finding a pattern that not only minimizes deblurring error with correct PSF but also maximizes deblurring error with incorrect PSFs. Accordingly, two objective functions are proposed. Both functions are defined in the frequency domain. A non-dominated sorting-based multi-objective evolutionary algorithm[22] is applied for finding a Pareto-optimal solution. An optimal pattern is chosen so that it can also discriminate before and after focal plane. As a result, an asymmetric pattern is proposed that is appropriate for depth estimation as well as deblurring in a single captured image.
According to[23], illumination condition and camera specification influence on the performance of coded aperture cameras. Therefore, our objective functions are formulated by considering the imaging circumstances. In this way, the designed mask has a reasonable throughput that ensures the captured image has an appropriate signal-to-noise ratio (SNR).

The proposed mask is compared with circular aperture, and some of state of the art coded aperture patterns. Performance comparison includes depth estimation accuracy as well as the quality of deblurring results.

In accord with the proposed objective functions, a depth estimation algorithm is introduced. In this method, a blurred image is deblurred with a set of PSFs. Then a PSF that provides the best quality deblurring result is selected as the correct blurring kernel. The quality of deblurred images is measured by an aggregate measure of no-reference image quality assessment criteria.

The rest of this paper is organized as follows: In Section 2, the problem is formulated and then, pattern evaluation functions are introduced. Section 3 describes the optimization method that is used to find the best pattern. Then, our depth estimation algorithm is presented in Section 4. Experimental results in both synthetic and real scenes are presented in Section 5. Finally, conclusions are drawn in Section 6.

## 2. Aperture Evaluation

In this section, first blurring problem is briefly reviewed. Then, our criteria for evaluating aperture patterns are introduced. Based on the proposed criteria, a multi-objective function is defined that is appropriate for comparing aperture patterns with different throughputs.

### 2.1. Problem Formulation

A binary coded mask with $n$ open cells can be imagined as a grid of size N×N, where $n$ number of holes distributed over the grid, are kept open [16, 23]. The pattern of open holes determines the shape of PSF, and the number of them specifies the mask throughput.

As stated in [23], an aperture pattern must be evaluated by consideration of both the shape and the throughput. Therefore, we redefine the well-known defocus problem with respect to these factors.
For a simple fronto-parallel object at depth $d$, defocusing is defined as convolution of a defocus kernel or PSF, called $k^d$ with a sharp image($f_n$) that causes spatial invariant blur:

$$f = k_d \otimes f_n + \omega_n,$$
$$\omega_n \sim N(0, \sigma_n^2) \quad , \quad \sum_i k_d^i = 1 \quad (1)$$

Equivalently, spatially invariant blur in the frequency domain is defined as Eq. 2:

$$F = K_d \,.\, F_n + \Omega_n \qquad (2)$$

The subscript $d$ indicates that the kernel $k$ is a function of depth of scene. The subscript $n$ declares that the brightness of the sharp image ($f_n$) and the amount of added noise ($\omega_n$) depend on the aperture throughput ($n$). Because of the additive properties of light, in a constant definite exposure time, the brightness of the sharp image ($f_n$) is increased linearly by increasing the number of open holes. The value of $\omega_n$ also changes by the number of holes. In this study, the growth of $\omega_n$ is studied by considering the number of holes, imaging system's specifications and scene illumination.

### 2.2. Noise Model

Imaging noise can be usually modeled as the sum of two distinct factors: read noise and photon noise [23]. Read noise is considered to be independent of the measured signal and is commonly modeled by a zero mean Gaussian random variable $r$ with variance $\sigma_r^2$. Photon noise on the other hand, is a signal dependent noise with Poisson distribution. When the mean value of photon noise is large enough, it is well approximated by a random Gaussian variable with equal mean and variance ($\mu = \sigma_p^2 = J_n$) [23],[24]. $J_n$ refers to the average number of photons received by each single pixel in a camera with an $n$ open-hole aperture.
As stated in [23], the total noise variance is computed as follows[23]:

$$\sigma_n^2 = \sigma_r^2 + \sigma_p^2 = \sigma_r^2 + J_n = \sigma_r^2 + n.J \qquad (3)$$

In this study, the average signal value in photoelectrons ($J$) of a single-hole aperture is computed by[23]:

$$J = 10^{15} . \frac{1}{F\#^2} . R . I . q . \Delta^2 . t \quad (4)$$

In our experiments, scene and imaging system parameters are assumed as follows which are typical settings in consumer photography:

| |
|---|
| q: sensor quantum efficiency = 0.5 (typical for CMOS sensors) |
| R: average scene reflectivity = 0.5 |
| t : exposure time = $10^{ms}$ |
| $\Delta^2$: pixel size = $5.1 \times 5.1^{\mu m}$ (SLR camera, typically Canon 1100D) |
| F# : aperture setting = 18 |
| I : scene illumination = 300 lux (typically office light) |

In the following section, first our criteria are proposed in terms of intensity level images. Next, we redefine the proposed formula in terms of photoelectron so that masks with different throughput can be compared.

### 2.3. Mask search Criteria

Suppose the image $F_n$ is blurred with an unknown Kernel $K_1$ (Eq.2). If it is deblurred with a typical kernel $K_2$ and Wiener filter is used for deconvolution, then total error of deblurring ($e_n$) is computed as Eq.5:

$$\mathrm{e_n} = F_\mathrm{n} - \hat{F}_n = \mathrm{F_n} - \frac{K_2^* F}{|K_2|^2 + |C_n|^2} = \mathrm{F_n} - \frac{K_2^*(K_1 F_n + \Omega_\mathrm{n})}{|K_2|^2 + |C_n|^2} = \frac{F_n K_2^* K_2 + F_n |C_n|^2 - K_2^* K_1 F_n - K_2^* \Omega_\mathrm{n}}{|K_2|^2 + |C_n|^2}$$

$$= \underbrace{\frac{F_n K_2^*(K_2 - K_1)}{|K_2|^2 + |C_n|^2}}_{e_n^{(1)}} + \underbrace{\frac{F_n |C_n|^2 - K_2^* \Omega_\mathrm{n}}{|K_2|^2 + |C_n|^2}}_{e_n^{(2)}} = e_n^{(1)} + e_n^{(2)} \qquad (5)$$

where $|C_n|^2$ is defined as the matrix of the expected value of noise to signal power ratios (NSR) of natural images. (i.e. $|C|^2 = \frac{\sigma^2}{\mathrm{A}}$ where A is the expected power spectrum of natural images and $\sigma^2$ is the variance of additive noise[18].) Eq. 5 shows total error consists of two parts:

$$e_n^{(1)} = \frac{F_n K_2^*(K_2 - K_1)}{|K_2|^2 + |C_n|^2} \quad , \qquad \text{error of wrong kernel estimation} \quad (6)$$

$$e_n^{(2)} = \frac{F_n |C_n|^2 - K_2^* \Omega_\mathrm{n}}{|K_2|^2 + |C_n|^2} \quad , \qquad \text{deblurring error} \quad (7)$$

If an accurate PSF is used for deblurring (i.e. $K_1 = K_2$), then the only term that determines total error of deblurring is $e_n^{(2)}$($i.e.\ e_n^{(1)} = 0$ ). On the other hand, by using a wrong kernel as PSF ($K_1$ # $K_2$), both $e_n^{(1)}$ and $e_n^{(2)}$cause some error in deblurring result. As we see in Section 3.2, the values of $e_n^{(1)}$ are much greater than $e_n^{(2)}$(See Figure 2). Therefore, when $K_1$ # $K_2$, $e_n^{(1)}$is the main determinant of the total error. Hence, according to our objective, a suitable pattern is defined as a pattern that minimizes the norm of $e_n^{(2)}$ as well as maximizing the norm of $e_n^{(1)}$. Norm of $e_n^{(1)}$ is computed as follows:

$$\left\| e_n^{(1)} \right\|_2^2 = \left( \frac{F_n K_2^*(K_2 - K_1)}{|K_2|^2 + |C_n|^2} \right)^* \left( \frac{F_n K_2^*(K_2 - K_1)}{|K_2|^2 + |C_n|^2} \right) = |F_n|^2 |K_2|^2 \frac{|K_2 - K_1|^2}{||K_2|^2 + |C_n|^2|^2} \qquad (8)$$

Since power spectra of all natural images follow a certain distribution, we can compute the expectation of $\left\| e_n^{(1)} \right\|_2^2$ with respect to $F_n$. According to *1/f* law of natural images[25], expectation of $|F_n|^2$ is computed as follows:

$$A_n(\xi) = \int_{F_n} |F_n(\xi)|^2 \, d\mu(F_n) \qquad (9)$$

where ξ is the frequency and $\mu(F_n)$ is the measure of sample $F_n$in the image space[18]. Accordingly, the expectation of $\left\| e_n^{(1)} \right\|_2^2$ is computed as Eq. 10.

$$D_n(K_2, K_1) = \mathbb{E}_{F_n}\{\left\| e_n^{(1)} \right\|_2^2\} = \sum_{\xi} \frac{A_{n_\xi} |\mathrm{K}_2|_\xi^2}{(|K_2|_\xi^2 + |C_n|_\xi^2)^2} |K_2 - K_1|_\xi^2 \qquad (10)$$

This measure can be considered as a distance criterion between two kernels. It can also help to make difference between defocus points lying front or back of the focal plane. We remind that the defocus PSF in front of the focal plane is the flipped version of the defocus PSF in the back of the focal plane (See Fig 1.a). It means these PSFs have a similar spectral response yet different phase properties. Eq. 10 includes the term $K_2$-$K_1$, which computes both spectral and phase differences of two kernels. Hence, by having an asymmetric aperture, deblurring with the flipped version of a PSF makes the error of $e_n^{(1)}$ and helps to distinguish the side of the focal plane (See fig 1.b).

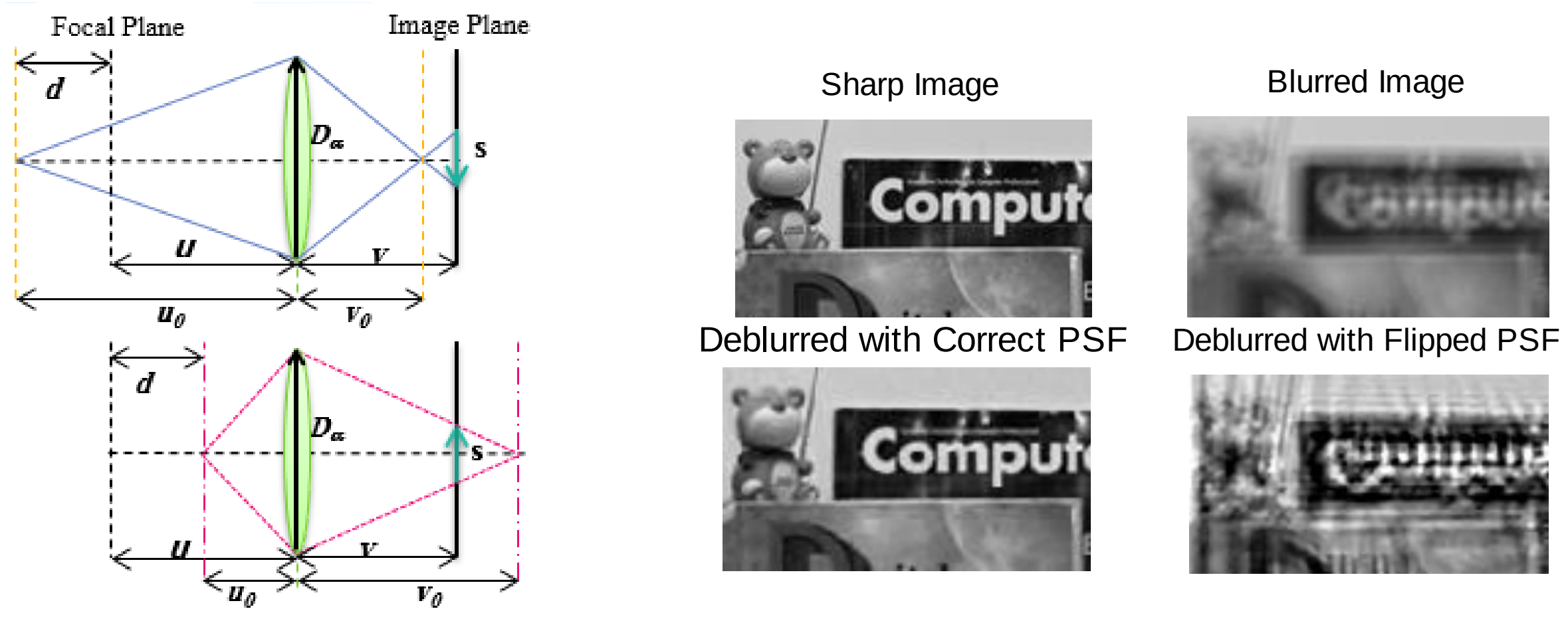

(a) (b)

Figure 1. (a)defocus PSF in front of the focal plane is the flipped version of defocus PSF in the back of focal plane. (b) If an asymmetric pattern is used for imaging,then deblurring with flipped PSF yields more error in the deblurred image (see Eq.10)

The expectation value of $\left\| e_n^{(2)} \right\|_2^2$ is computed in a similar manner. (Details are found in ref.[18]):

$$R_n(K_1) = \left\| e_n^{(2)} \right\|_2^2 = \sum_{\xi} \frac{\sigma_n^2}{|K_1|_\xi^2 + |C_n|_\xi^2} \quad (11)$$

This value has been used by Zhou et al.[18] as a metric to find aperture patterns that yield less error in deblurring results. However, here we redefine it to be able to study patterns with different throughputs. In addition, we search for a pattern that is suitable for depth estimation as well as deblurring.

If the camera response function[26] is assumed to be linear, then relations 10 and 11 can be stated in terms of photon as follows:

$$\begin{cases} D_n(K_2, K_1) = \sum_{\xi} \dfrac{{J_n}^2 . {A_1}_\xi |\mathrm{K}_2|_\xi^2}{\left(|K_2|_\xi^2 + \frac{\sigma_r^2 + J_n}{{J_n}^2 . {A_1}_\xi}\right)^2} |K_2 - K_1|_\xi^2 \\ R_n(K_1) = \sum_{\xi} \dfrac{\sigma_r^2 + J_n}{|K_1|_\xi^2 + \frac{\sigma_r^2 + J_n}{{J_n}^2 . {A_1}_\xi}} \end{cases} \quad (12)$$

where $A_1$ refers to the expected power spectra of natural images taken with a single hole aperture. If we want to study patterns with different throughputs, then $D_n$ and $R_n$ must be normalized. In conclusion, our multi-objective function is defined as follows:

$$\begin{cases} min \quad R(K_{s_1}) = \dfrac{1}{n^2} . R_n(K_{s_1}) \\ max \quad D(K_{s_1}, K_{s_2}) = \dfrac{1}{n^2} D_n(K_{s_1}, K_{s_2}), \quad s_1 \neq s_2 \end{cases} , s_1, s_2 \in S \; and \; n \in [1..N^2]$$

$$s.to: \quad 0 \leq |K_s(\xi)| \leq 1 , \quad s \in S \quad (13)$$

where $S$ refers to a limited range of blur scales.

## 3. Aperture Pattern Design

### 3.1. Mask Resolution

In this study, mask resolution is determined such that each single hole provides no diffraction. On the basis of superposition property in coded aperture imaging, if a single hole of a pattern does not provide any diffraction, then the image composed of rays passed through all the holes does not provide any diffraction [27]. Based on the formula proposed in [28], a 7×7 mask is appropriate for an imaging system with aperture-diameter = $20^{mm}$ and pixel-size = $5.1^{\mu m}$. According to the camera specification used in our experiments, this resolution is selected for our mask and thus, the number of open holes (n) is in[1..49].

### 3.2. Optimization

Multi-objective optimization is usually described in terms of minimizing a set of functions. Therefore, we rewrite our objective functions as follows:

$$min \ \{R(K_{s_1}), -D(K_{s_1}, K_{s_2})\}, \quad for \ s_1, s_2 \in [1..10] \ and \ s_1 \neq s_2 \qquad (14)$$

Although these evaluation functions are clear and concise, solving them in the frequency domain is a challenging problem. Since we search for a binary pattern with specific resolution, the objective function must also satisfy some other physical constraints in the spatial domain. Deriving an optimal solution that satisfies all constraints in both frequency and spatial domain is difficult. Therefore, a heuristic search method is used for solving the problem. In evaluating each pattern, $R$ and $D$ values are computed for different size of kernels. Then, the maximum value of $R$ and minimum value of $D$ are used for evaluating the pattern.

The main goal of a multi-objective optimization problem is finding the best Pareto optimal set of solutions[22]. Here, the notion of Pareto optimal must be defined. In a multi-objective optimization problem that consists of $m$ functions $min\{z_1(x), \ldots z_m(x)\}$, a feasible solution $x$ dominates another feasible solution $y$, if and only if $z_i(x) \leq z_i(y)$ for $i$=1..$m$ and $z_j(x) < z_j(y)$ for at least one objective function $j \epsilon$[1..m]. A solution is called Pareto optimal if it is not dominated by any other solution in the solution space [22].

Among the heuristic search methods, Genetic Algorithms (GA) are appropriate to solve multi-objective optimization problems. A single-objective GA can be easily modified to find a set of multiple non-dominated solutions in a single run[22].Fast Non-dominated Sorting Genetic Algorithm (NSGA-II)[29] is a well-suited method for solving our problem. It uses Pareto-ranking approach that uses the concept of Pareto dominance in evaluating fitness or assigning selection probability to each solution. The chromosomes are ranked based on a dominance rule, and then a fitness value is assigned to each solution based on its rank in the population, not its actual objective function value. Furthermore, NSGA-II uses crowding distance with aim to acquire a uniform propagation of solutions along the best known Pareto front without using a fitness sharing parameter[22].

In this study, our problem is solved by NSGA-II[29]. A generation of binary patterns with population size 1500 is created. A pattern is defined by a vector of 49 binary elements. According to[30], this size of population is enough to converge to a proper solution. Other parameters are set by default values which have been adjusted in the prepared software[1]. Figure 2 shows the resulted values of objective functions in Pareto-front. The values of the proposed objective functions are also computed for some other apertures and then added to the figure.

[1] http://www.iitk.ac.in/kangal/

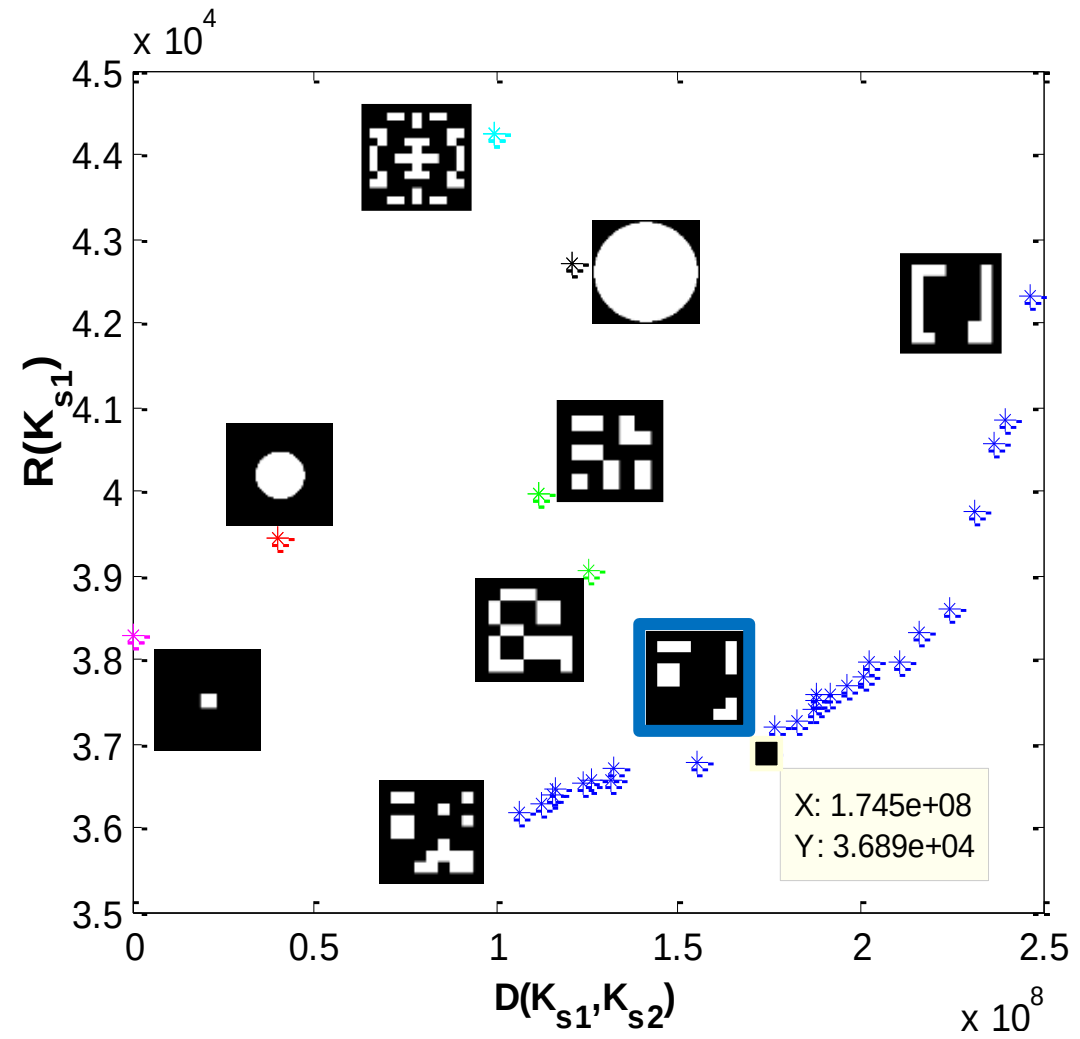

Figure 2 *D* values vs. *R* values of final patterns in the Pareto optimal solution (blue), Open circular aperture (black), conventional aperture (red), pinhole aperture(magenta), patterns proposed in[14] (green) and[15] (cyan). Final selected pattern has been highlighted by the blue border.

According to Figure 2, in the Pareto optimal solution, by increasing the symmetry in patterns, deblurring error (*R*) increases, as well as increasing the error of using wrong scale kernel (*D*). However, it doesn't mean that any symmetric pattern has better discrimination ability of kernel estimation. For example, objective functions are also computed for pinhole aperture, open circular aperture, circular aperture whose throughput is equal to the selected coded pattern (highlighted by blue border)[2] and the symmetric pattern proposed in[15]. Although these patterns are symmetric, they cannot provide larger *D* values than some asymmetric ones. On the other hand, all asymmetric patterns cannot provide smaller *R* values than any symmetric patterns. Indeed, the amount of *R* and *D* depend on several factors such as mask throughput and spectral properties. For example, one of the asymmetric patterns proposed in[14] results in more deblurring error than conventional aperture. Another example is pinhole aperture that makes no blur, but because of low throughput, captured image has low SNR and thus deblurring error is high. On the other hand, the open circular aperture has high throughput yet low Depth of Field (DoF).Therefore, blurring causes to drop lots of frequency components of the image, which yields low quality deblurring results.

As stated earlier, NSGA-II provides a set of solutions. Since, just one pattern has to be selected, we compute $D_r = D(K_{s1}, rot(K_{s1}, 180))$ for all the obtained patterns in the Pareto optimal solution. In a similar manner, this value is computed for asymmetric patterns proposed in[14]. Figure 3 shows the computed values.

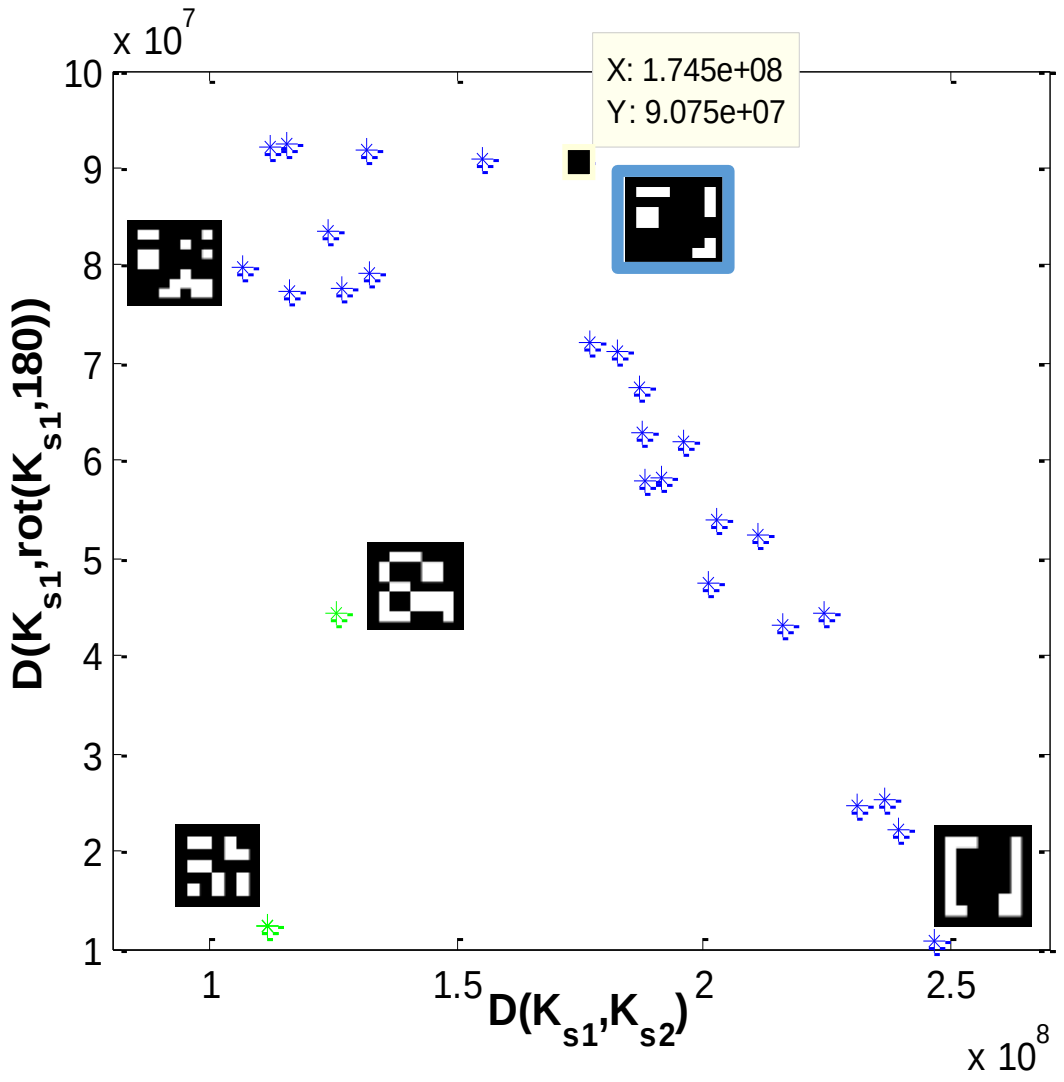

Figure 3 The value of *D* of wrong scale kernels vs. $D_r$ of the flipped correct scale for the obtained patterns by NSGA-II (blue) and asymmetric patterns proposed in[14] (green).

[2] In the rest of text, the circular aperture with the same throughput of selected coded pattern is called conventional aperture.

As shown in Figure 3, by increasing symmetry, $D_r$ decreases. According to the significance of the criterion $D_r$, the pattern highlighted with the blue border is selected as a sample of the proposed patterns. It must be mentioned that the selected pattern is not the best choice for all situations. However, since this pattern provides appropriate values of $D$, $R$ and $D_r$, it is selected as the final pattern. Indeed, the final pattern should provide a minimum value of weighted sum of all criteria, which each weight represents the importance of the related criterion. In this study, we use NSGA-II, which does not use the weighted sum for optimization.

In the first step of analyzing the selected pattern, its spectral properties are compared with the conventional aperture. We remind that both apertures have the same throughput. Therefore, in different imaging conditions, the same amount of additive noise are added to the captured images. In this situation, the spectral properties of the apertures determine the result. Figure 4 shows 1D slices of spectral response of these apertures at 5 different blur scales. Based on [15], if a pattern has a different frequency response in each scale, then the distinction of blur scales will be easier. As shown in Figure 4, for the conventional aperture, the zero amplitude in different scales overlap in some frequencies, which makes it hard to distinguish between the blur scales. However, the coded pattern has different spectral responses in different scales.

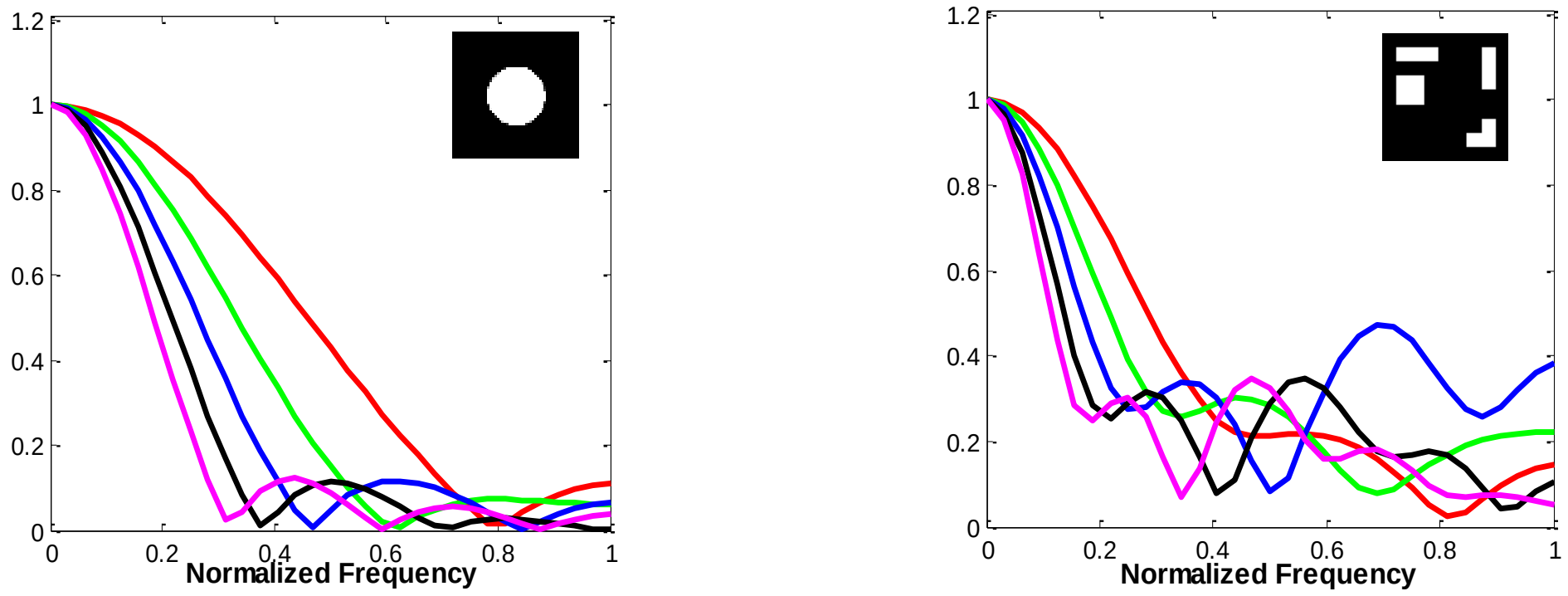

Figure 4 The 1D slide of spectral response at 5 different blur scales for conventional and coded aperture.

The spectral response of two studied apertures is also compared with together in 4 different scales. As shown in Figure 5, the minimum spectral response of our pattern is higher than the conventional aperture, especially in larger blur scales. Therefore, in the proposed pattern, fewer frequencies of the captured image are attenuated and thus deblurring result will have better quality.

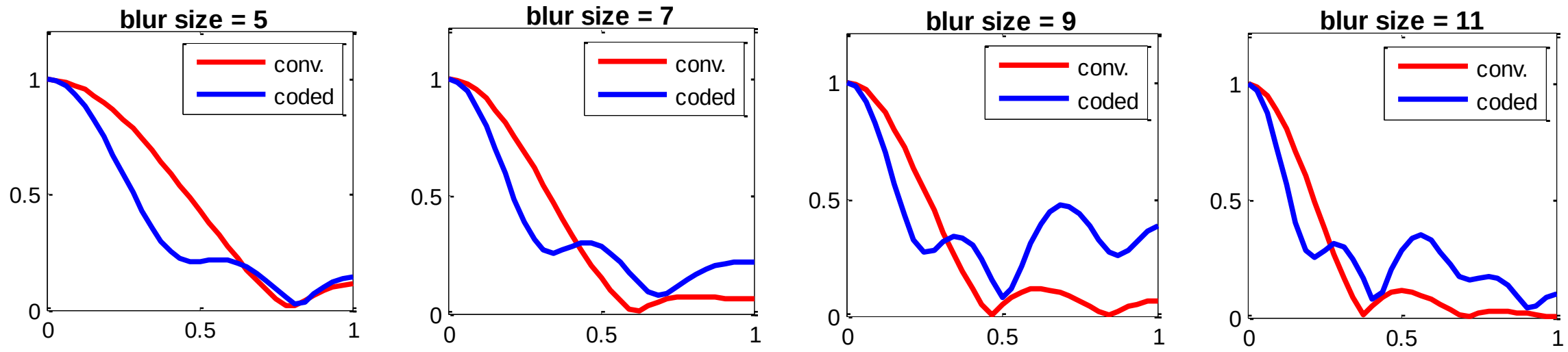

Figure 5 1D slices of Fourier transforms of conventional aperture (red) and the proposed pattern (blue) at 4 different scales.

Another advantage of the proposed pattern is its high sensitivity to the depth variation. It is known that DoF decreases by increasing the aperture diameter. In the proposed pattern, open holes are in the margin of the mask. Hence, this aperture pattern is more sensitive to depth variations than the conventional aperture. For studying the difference of depth sensitivity in these apertures, the size of blur is computed in a limited range of depth (before and after focal point) for a typical lens (EF 50mm f/1.8 II). Size of blur (s) is computed based on thin lens formula[27]:

$$s = \frac{D_a(v - v_0)}{v_0} \quad , \quad v_0 = \frac{F u_0}{u_0 - F} \quad , \quad v = \frac{F u}{u - F} \qquad (15)$$

The parameters used in Eq. 15 were introduced in Figure 1(a). The aperture diameter ($D_a$) is assumed to be $20^{mm}$ and $8.21^{mm}$ for coded and conventional patterns, respectively.

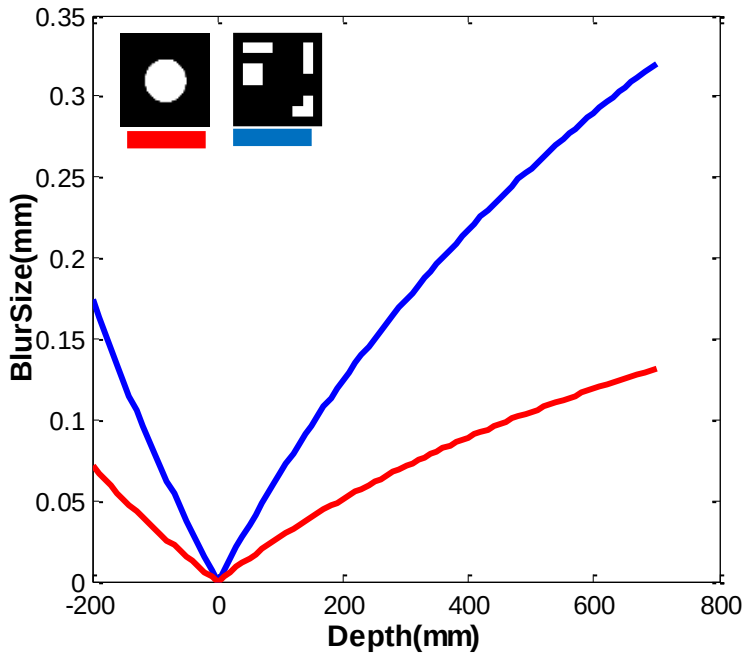

Figure 6 Blur size vs. depth for the conventional (red) and the coded (blue) apertures. (focus length (u) = 1200mm , v = 50mm). Code aperture is more sensitive to depth variation.

As shown in Figure 6, the proposed pattern is more sensitive to depth variation. Therefore, depth estimation is easier in the images captured by the coded pattern. On the other hand, according to Figure 5, coded mask gives a higher spectral response. Hence, it is expected to obtain better results in both deblurring and depth estimation in real imaging.

## 4. Depth Estimation

In this section, a new method for depth estimation is described. Several methods have been proposed for estimating depth from defocus. Levin et al.[15] compute reconstruction error of using different scales of PSF for deblurring. A PSF that yields minimum error is chosen as the correct scale of PSF.

Martinello et al.[31] show all natural images blurred with a scale *s*, are mapped to a specific subspace. A learning based approach is used to find basis vectors of each subspace that corresponds a blur scale. For a region with constant depth, the distance of each subspace to this region is computed. A subspace with the nearest distance determines the depth of that region. Since the linear subspaces for a kernel and its flipped version are the same, this method cannot distinguish between blurred scenes which are before and after focal plane[14, 27]. Increasing the amount of noise or size of blur reduces the distance between subspaces. Therefore, the precision of this method reduces in noisy or highly blurred images [16]. Sellent et al.[14] use this method to determine the scale of PSF. Then, they use a quality assessment based method to find the direction of PSF. Our proposed method is almost similar to[14]. However, in our method, no prepared database is used for PSF estimation. It is based on the proposed objective function (Eq.13) which can be used for detecting both scale and orientation of PSFs. As stated earlier, deblurring with a wrong kernel yields a low quality image. Figures 7 shows an image blurred with a kernel of conventional aperture and then deblurred with different scales of the kernel. The quality of each image is computed with a hybrid no-reference quality measure proposed in[32]. Both visual results and quality measures show that the best quality is obtained for the deblurred image with correct scale (i.e. r=3). Deblurring with smaller kernels yields blurry images and deblurring with larger PSFs yields images with artifacts.

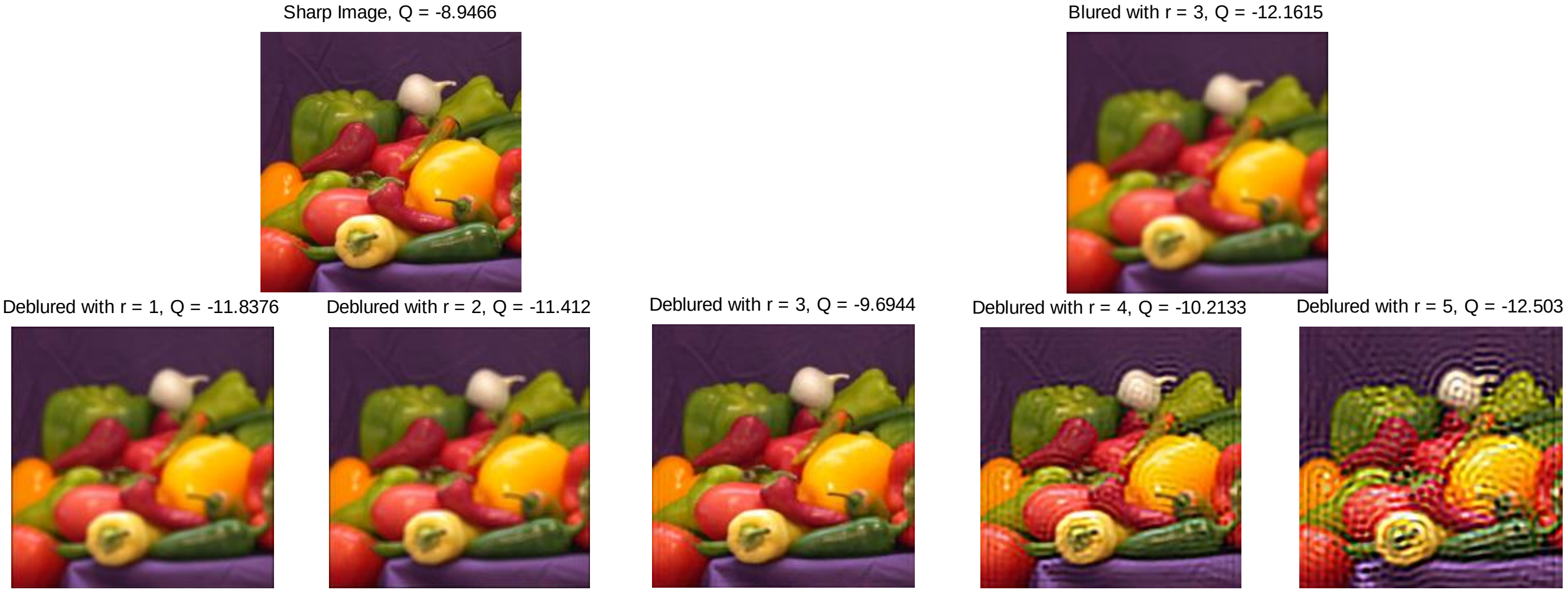

Figure 7 Deblurring results with different radii of the kernel(r=1..5) in imaging with conventional aperture. The quality of each image is computed by the no-reference quality assessment measure proposed in[32]. A larger Q-value means better quality.

This experiment is also repeated for the proposed asymmetric pattern. Figure 8 shows deblurred images with 5 different blur scales and their flipped versions. It shows, deblurring with wrong kernels yields low-quality images while deblurring with the correct kernel yields a high quality image.

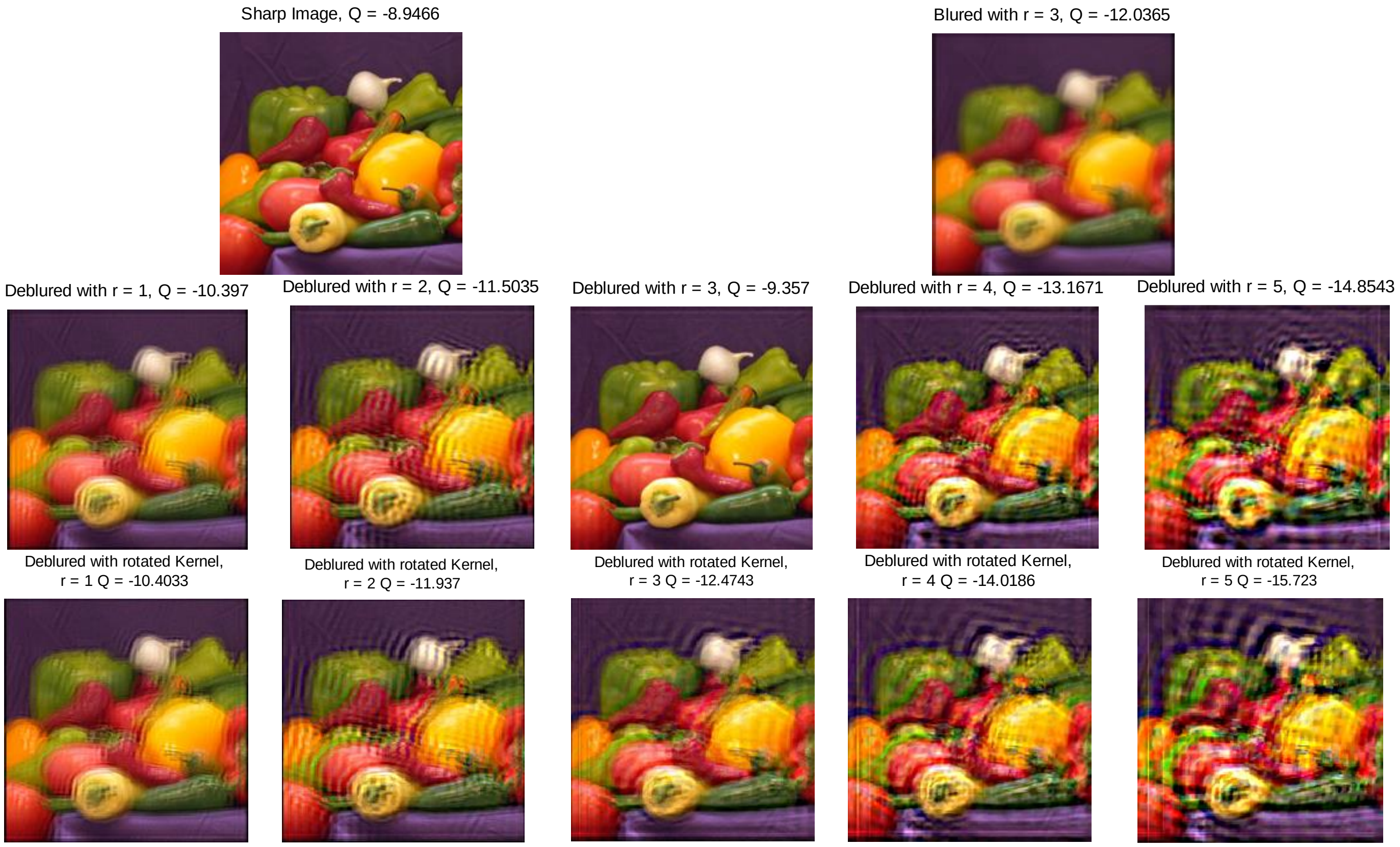

Figure 8 Deblurring results of coded aperture imaging with different size and rotation of kernels(r=1..5). The quality of each image is computed by the no-reference quality assessment measure proposed in[32]. A larger Q-value means better quality.

As shown in Figures 7 and 8, deblurring with wrong kernels (in size or orientation) produces low-quality images. On the other hand, deblurring with correct scale (especially in our mask) gives a high-quality image. As a result, among a limited range of blur scales, we can estimate blur kernel by deblurring image with each kernel and compute the quality of the restored image. The best quality determines the correct kernel. Up to now, several measures have been proposed as no-reference image quality criteria. One of the most comprehensive studies[32] uses weighted sum of 8 different criteria for evaluating the quality of an image. (Recent studies show using an aggregate measure of image quality assessment criteria is more accurate [19, 32]) Although this measure is applicable for depth estimation, it is more complicated than it is needed. In our application, the quality of deblurred versions of the same image is compared with each other. Indeed, the quality measure is used as a relative measure not a strict measure. Therefore, much easier measures are applicable for quality assessment. Reducing the number of criteria improves the speed of depth estimation algorithm. In this study, the quality of deblurred imaged are evaluated by an aggregated measure containing four criteria, which work best for our application. These criteria are sensitive to blur or artifact or both of them.

- *Norm Sparsity Measure[33]*

This criterion is defined as the ratio of the $l_1$ norm to the $l_2$ norm of the high frequencies (gradient) of an image, and was originally used as a regularization term in blind deconvolution. Krishnan et al.[33] show both blur and noise cause increasing this measure. It means the lowest cost of this measure belongs to the original undistorted image. Our experiments showed this simple yet effective measure can determine the best quality image in a set of deblurred images. However, when it is used for depth estimation in small patches of the image, some errors occur in determining the best quality patch.

- *Sparsity priors[15]*

On the basis of this measure, gradients of natural images follow a heavy-tailed distribution that can be defined by $|\nabla f_0|^{0.8}$ ($f_0$ refers to the original image). Deblurring with wrong kernels increases this value. This measure is also used by Levin et al.[15] as a regularization term for deblurring.

- *Sharpness Index[34]*

Sharpness index is a quality assessment measure that is also used as a regularization term in blind deconvolution. It is sensitive to both blur and artifact and is computed on local window of images via $SI(w) = -\log_{10}\Phi(\frac{\mu - TV(w)}{\sigma})$ where TV refers to total variation[3] of window $w, \sigma^2 = var(TV(w)$ and $\mu = \mathbb{E}(TV(w))$. The function $\Phi(x) = (2\pi)^{-0.5}\int_x^{+\infty} e^{-t^2/2}dt$ is used for computing the tail of the Gaussian distribution [34].

- *Pyramid Ring[32]*

Aggregating of three mentioned measures makes a powerful measure for detecting the best quality image. However, in very few small patches of images with special textures, there are yet some errors in kernel detection. Therefore, a measure that uses both blur image and deblurred image for ringing detection is also used as the fourth measure. Pyramid Ring estimates the amount of ringing artifacts in the deblurred image by comparing the gradient map of blur and deblurred image.

The quality assessment aggregate measure is defined as the weighted sum of the four mentioned criteria. The weights are computed by statistical regression methods. For computing weights, 40 image patches with different textures and edges are selected. They are blurred with 21 kernels in size (-11:+11) and then deblurred. Since in this step, reference patches exist, the quality of each deblurred patch (called test patches) is computed by a full-reference quality assessment measure. Then, using logistic regression[35], the weights of 4 criteria are computed in a manner that their weighted sum is proportional to the quality value obtained by the full-reference measure. The full-reference measure used in this study is a measure proposed in[19]. It is an aggregate measure consists of RMSE, SSIM and HDR-VDP2 which is defined as $((1 - RMSE) + SSIM + \frac{HDR_VDP2}{100})$. (Images are assumed to be gray level [0..1] and thus the value of each term is in the range of [0..1]). Power and accuracy of these measures have been studied in[19]and[32].

Finally, the aggregate quality measure is defined as follows:

$$Quality = -12.65 * normSps + 0.073 * sharpIndex - 0.289 * sparsity - 9.86 * pyrRing \quad (16)$$

In this measure, a higher value means more quality. We use this measure to evaluate the quality of deblurred images (or patch of images) obtained by different PSFs. A PSF, which yields a deblurred image with the best quality is chosen as the right PSF. This method is used for detecting both size and direction of the PSF.

It must be mentioned that training steps were also repeated for other studied aperture patterns. However, the weights did not change significantly.

### 4.1. Handling Depth Variations

Real world scenes include depth variation. Therefore, each part of an image may be blurred with a different kernel. A common method for depth estimation in these images is using almost small patches, within each patch the depth is assumed to be constant. Blur kernel is estimated for the patch, and this estimation is assigned to the central pixel of the patch. By repeating this stage for all pixels of the image a raw depth map is attained. Then coherent map labeling is performed by using the raw depth map, image derivative information and some smoothness priors [6, 15].

---

[3] $TV(v) = \|\partial_x(v)\|_1 + \|\partial_y(v)\|_1$

In this study, at the first stage, 2 blur scales that give deblurred patches with the highest quality are considered as the possible true scales of the central pixel. The probability of each scale is computed based on their relative computed quality. More quality yields more probability and the sum of two probability values are equal to 1. Zhu et al.[6] use three possible blur scales and color information for depth map estimation. We found by experiment that using two probable values is enough to improve the final depth map. At the end of this stage, a three-dimensional matrix is obtained. In other words, for a H×W image and S possible depths, matrix $D_R \epsilon \mathbb{R}^{H\times W\times S}$ includes the raw depth map in which $D_R(h,w,s)$ represents the probability of occurring depth $s \in S$ in pixel (*h*,*w*).(In the rest of text, for simplicity, the position of each pixel is noted by single symbols like *p*, *q*.)

The raw depth map may have some errors in the depth estimation, especially in depth discontinuities. Therefore, in the second step, a coherent blur map is attained by minimizing an energy function that is used in image segmentation approaches. This function is defined as follows[36]:

$$E(D_c) = \sum_{p} D_p(s_p) + \sum_{(p,q)\in N} \lambda_{p,\mathrm{q}} V(s_p, s_q) \qquad (17)$$

where *p* and *q* refer to image pixels. The first term $D_p(s_p)$ reflects fidelity to the previous probability blur scale (*s*) estimation at position *p*. The second term $V(s_p, s_q)$ is a smoothness term, which guarantees neighbor pixels with similar gray levels have similar blur scales. $D_c$ denotes a solution for coherent data map whose energy (*E*) is the minimum.

We use the method proposed in [6] for coherent map estimation. Here, we briefly review this method. To assign a penalty to depth change in $D_p$, the early probabilities of blur scale ($p_p(s)$) are convolved with a Gaussian filter ($N$ (0, 0.1)) to reach the smoothed probabilities ($\hat{p}_p(s)$). Then $-\log(\hat{p}_p(s))$ is used as $D_p(s)$. (See ref.[6]). This function could also be used for cases that one or more probabilities are assigned to the initial blur scale.

The smoothness term $V(s_p, s_q)$ examines depth discontinuity in neighboring pixels. For each pixel *p*, depth similarity is checked with its 8 surrounding pixels with the following equation:

$$V(s_p, s_q) = |s_p - s_q| \qquad (18)$$

The relative importance of difference between depths of two adjacent pixels is determined by the difference of their gray level ($g_p$ and $g_q$). Hence, $\lambda_{x,\mathrm{y}}$ is defined as follows[6]:

$$\lambda_{p,q} = \lambda_0 \exp\left(-\frac{\|g_p - g_q\|^2}{\sigma_\lambda^2}\right) \qquad (19)$$

In our experiment, parameters are set to $\lambda_0$=1000 and $\sigma_\lambda$=0.006. Finally, α-expansion is used to minimize the energy function [37].

## 5. Experiment

The proposed mask and depth estimation method are validated in several experiments. It is compared with circular aperture, conventional aperture and two other masks designed for depth estimation[14, 15]. Among the masks proposed by Sellent et al.[14], we choose the 7×7 mask that is the best one based on our evaluating criteria (see Fig. 2 and 3). Our study contains synthetic and real experiments. It is expected the designed mask helps to exact PSF estimation and then provides good deblurring results.

### 5.1. Synthetic Experiments

*A. Experiment 1*

In the first experiment, a number of various images are blurred uniformly with various blur scales (*s*=1:14). Then 60 patches of these images are randomly selected and their depth is estimated by using the method described in Section

4. Figure 9(a) shows some of the selected patches. In each scale, the mean and variance of estimated size of PSFs are computed over all patches. This experiment is repeated for different aperture patterns and in three levels of noise (σ = 0.001, 0.005, 0.01). Based on the results shown in Figure 9(c), increasing noise decreases depth estimation accuracy. However, results are almost appropriate especially in our mask and the mask proposed by Sellent et al.[14]. It must be mentioned since in this experiment both symmetric and asymmetric patterns are studied, only one side of the focal plane is considered.

To have better comparison among studied aperture patterns, in each scale, the norm of difference between the ground truth blur scale ($s_{gt}$) and the estimated blur scale ($s_{es}$) is computed over all patches (i.e.$\sum_{p=1}^{50} \left(s_{gt}^{p} - s_{es}^{p}\right)^2$). Then this value is averaged over all studied blur scales[14]. Figure 9(b) shows mean square error (MSE) of depth estimation for different apertures at three amounts of noise. It shows in equal circumstances where all imaging conditions (including throughput) are the same; coded pattern performs better than its corresponding conventional aperture.

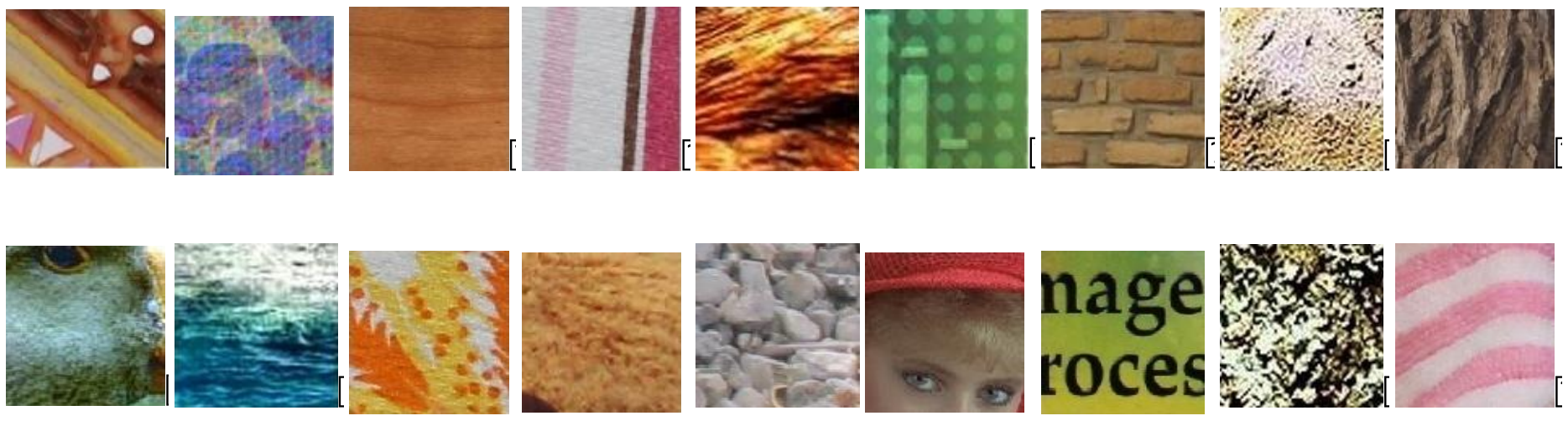

(a) A few number of patches used in the experiments

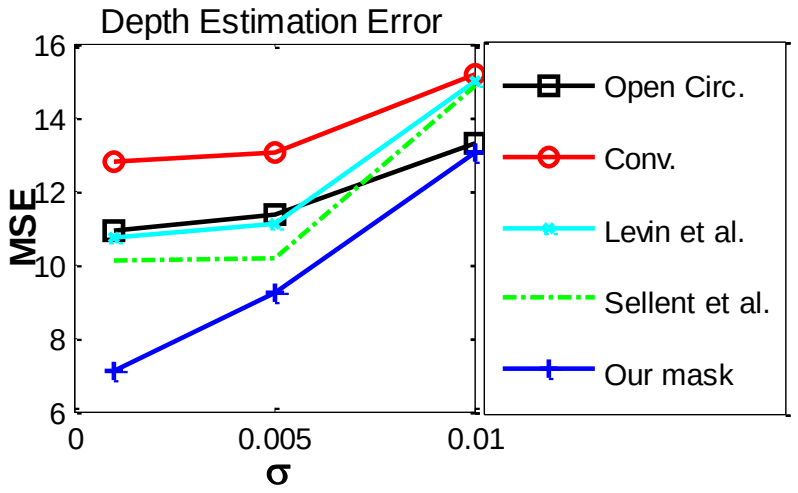

(b) Average of depth estimation error in 14 sizes of blur (s = 1:14).

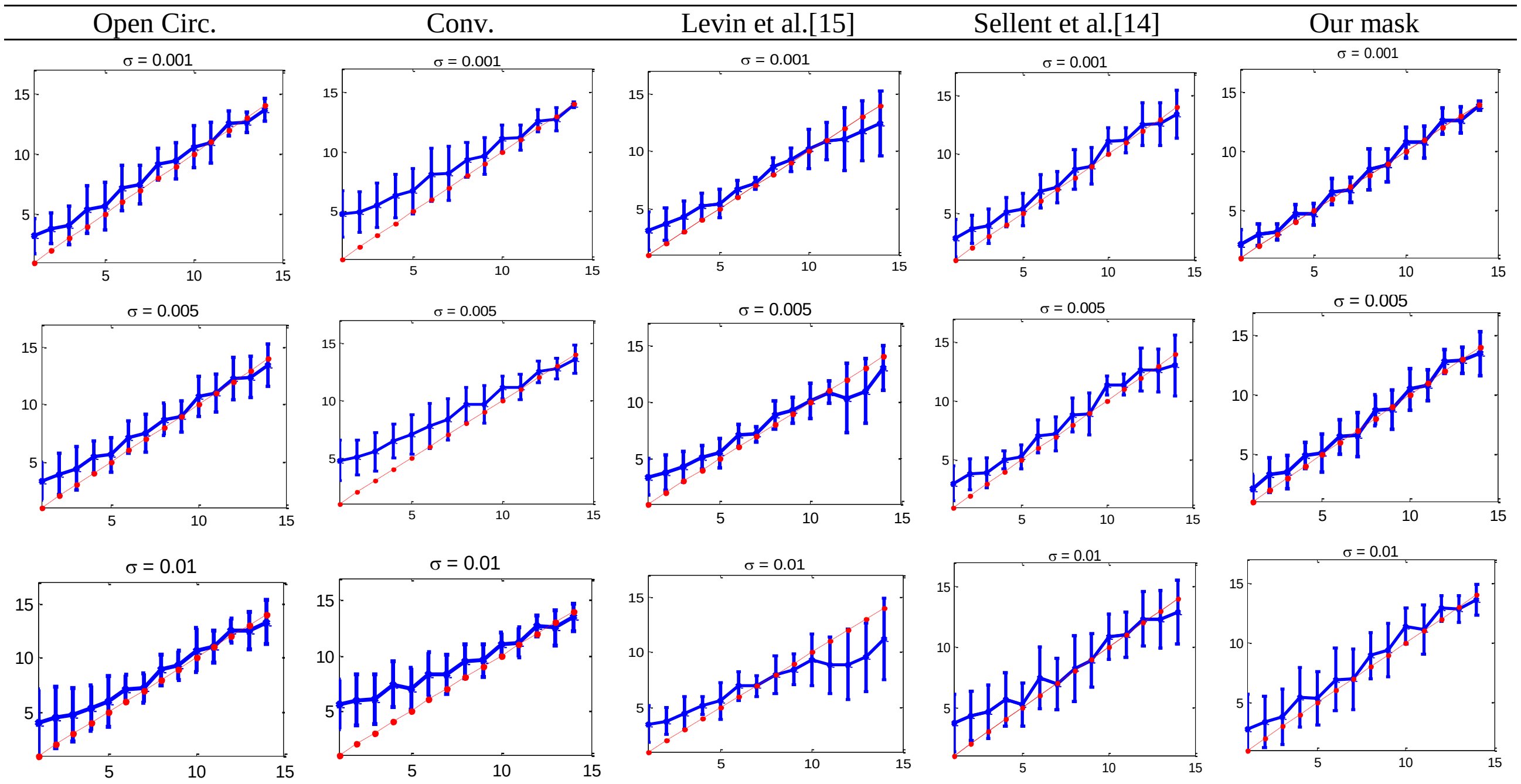

(c) Average and variance of estimated blur scale in comparison with ground truth scale ( red diagonal).

Figure 9 Results of depth estimation for five apertures at 3 noise levels (σ=0.001,0.005,0.01) and 14 blur sizes (s = 1:14).

Depth estimation experiment is repeated for asymmetric patterns with blur sizes in the range from -12:+12 pixel. Since a blur size of 0 is meaningless and, ±1 corresponds to the sharp image, 23 different sizes of blur are indeed examined. Figure 10 shows our method provides appropriate results at σ = (0.001, 0.005) and depth estimation error (MSE) of the proposed aperture is less than the pattern in[14].

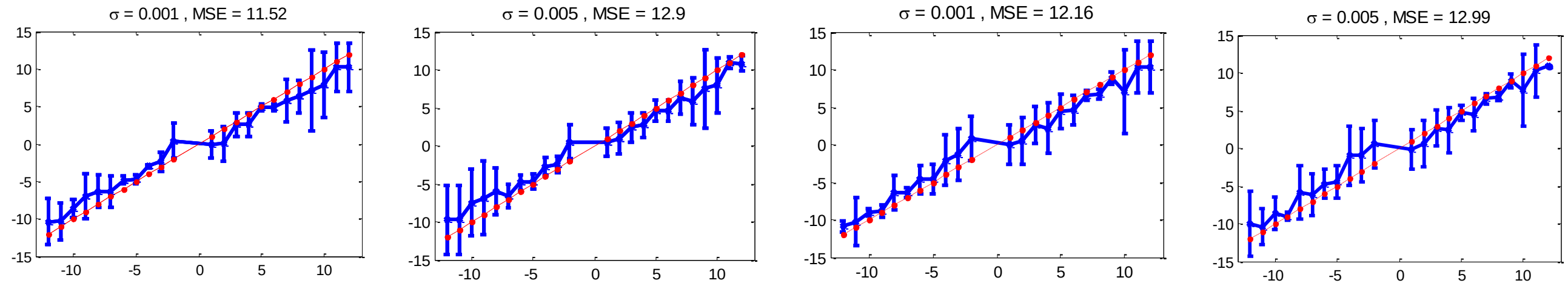

(a) Our mask (b) Mask proposed by Sellent et al.[14]

Figure 10 Average and variance of estimated blur scale in comparison with ground truth scale (red diagonal) at 2 noise levels(σ=0.001, 0.005) in the depth range -12:12.

### B. *Experiment 2*

In the second experiment, deblurring results of aperture patterns are studied. For different scales of blur, each of the studied blurred patches is deblurred with correct scale of PSF. Then, Root Mean Square Error (RMSE) of difference between original sharp image and its deblurred version is computed. The average of RMSE over all patches is computed. As shown in Figure 11, our pattern provides the least error especially in large blur scales while the conventional aperture is the best aperture in lower blur scales. A sample of deblurring result of Circular Zone Plate (CZP) chart is shown in Figure 12. In all experiments images are deblurred by the well-known sparse deconvolution algorithm[4] proposed by Levin et al.[15].

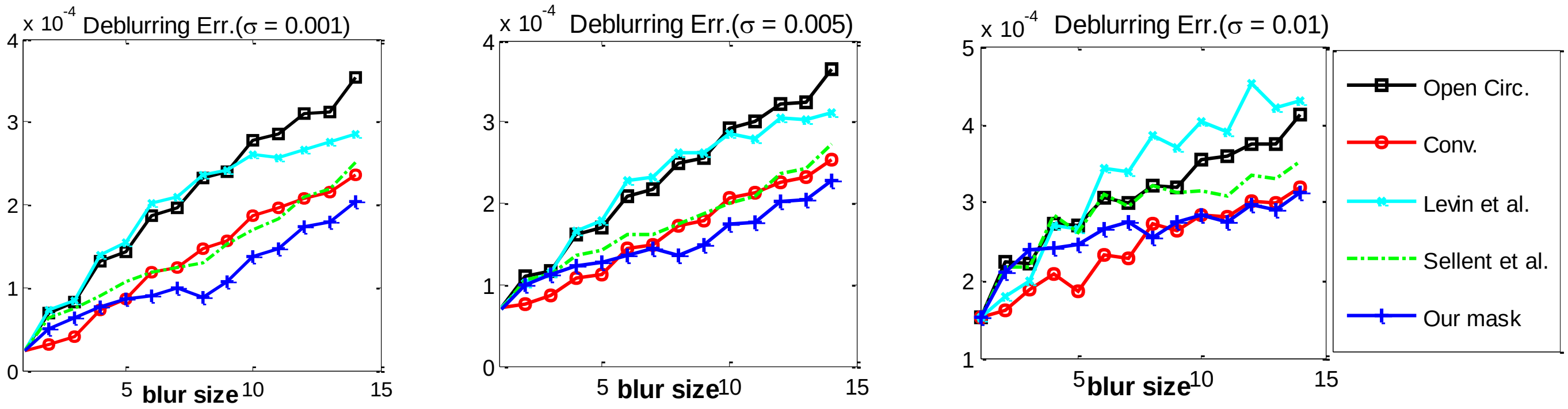

Figure 11 Deblurring error of five apertures at 3 noise levels (σ=0.001, 0.005, 0.01) for 14 blur sizes (s = 1:14).

[4] http://groups.csail.mit.edu/graphics/CodedAperture

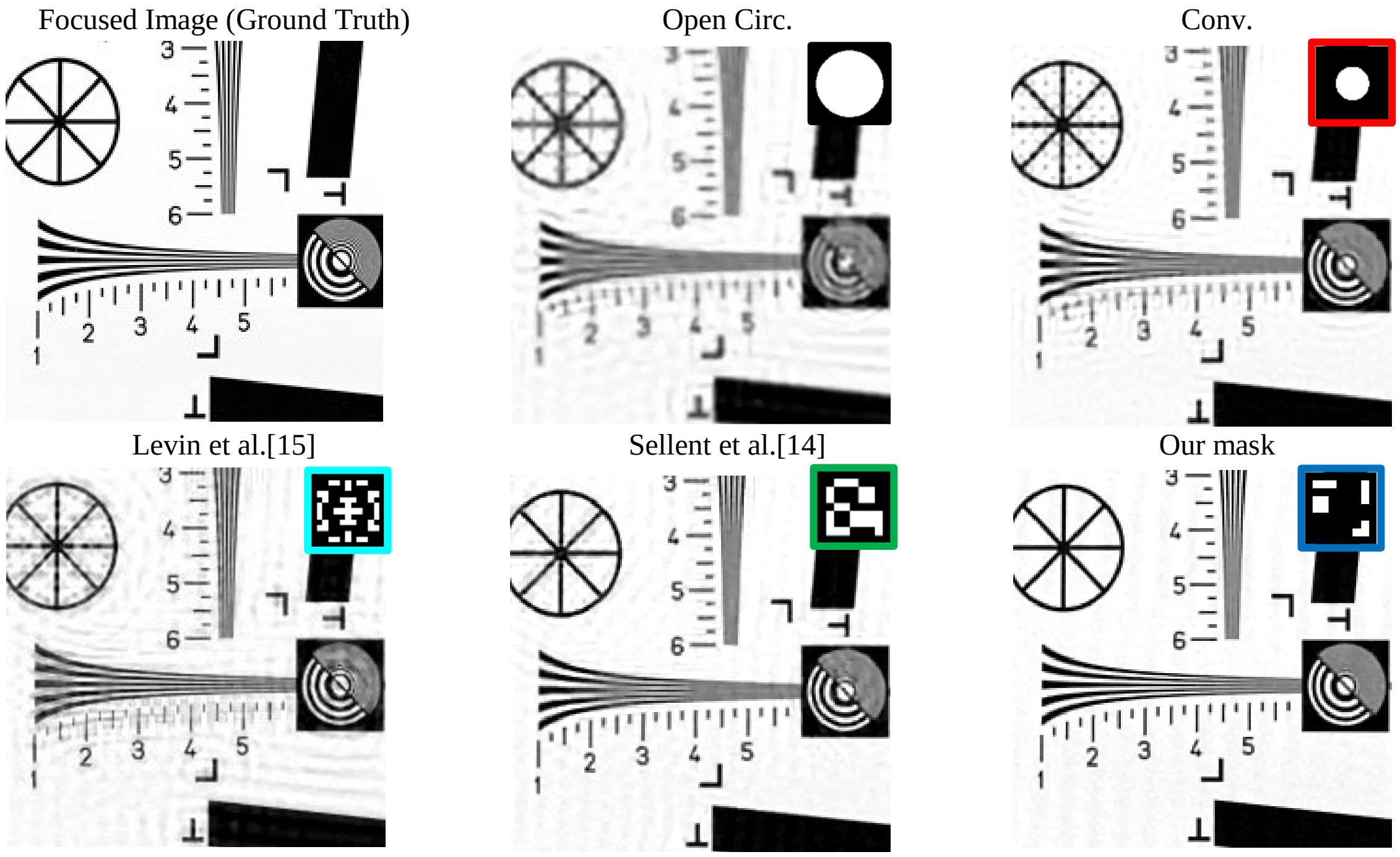

Figure 12 Comparison of deblurring results obtained using different aperture patterns (blur size = 13, σ=0.005)

### 5.2. Real Scene

For real experiments, the proposed pattern is printed on a single photomask sheet. It is cut out of the photomask sheet and inserted into a camera lens. In our experiment, a Canon EOS 1100D camera with an EF 50mm f/1.8 II lens is used. The disassembled lens and assembled with the proposed mask are shown in Figure 13(a, b).

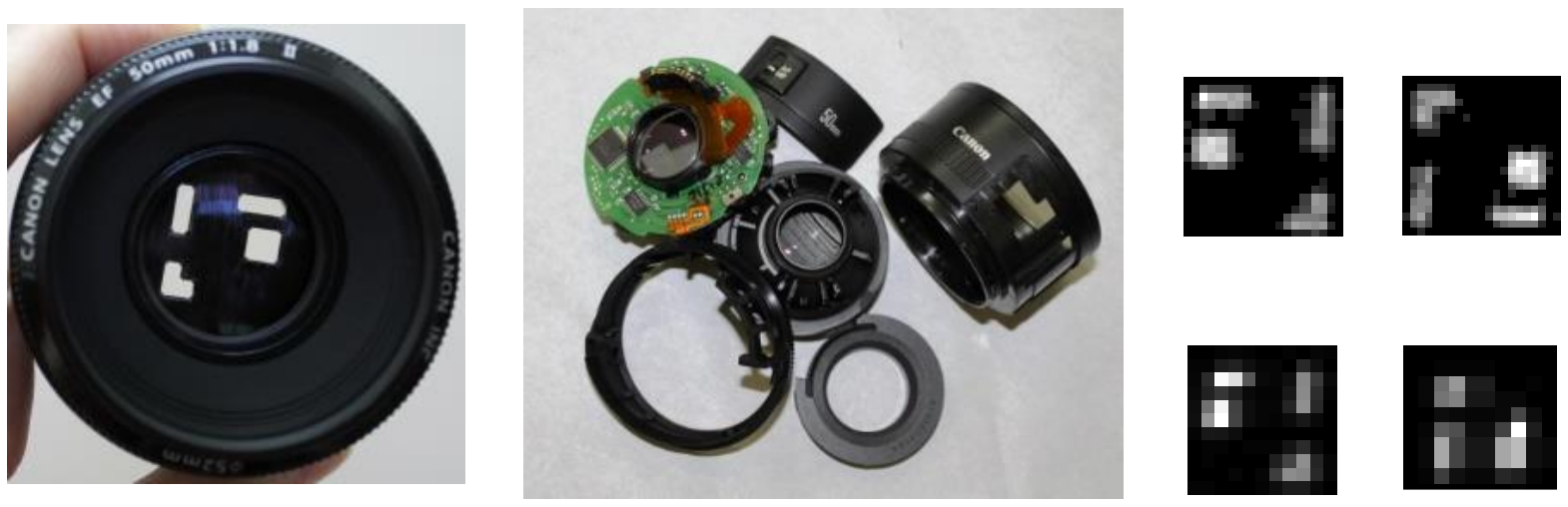

(a) (b) (c) (d)

Figure 13 (a) lens assembled with the proposed mask, (b) disassembled lens. (c), (d) calibrated PSFs of evaluated pattern

A very thin LED is used to calibrate the true PSF. The LED is mounted behind a pierced black cardboard to make a point light source. Since, in each experiment, the position of focal point may be changed, the camera focus is set to a sample point, then, the camera is moved back and forth up to $60^{cm}$ in $5^{cm}$ increments. At each depth, an image is captured. Each image is cropped according to the surface that the point light spreads. Afterward, by using some threshold values, the residual light is cleared and the result is normalized. In some rare cases, there is a jump in PSF scale in consecutive measured PSFs. In these cases, other PSF scales are generated synthetically from the obtained PSFs. In this way, a bank of PSFs is generated that covers all possible sizes of PSF in the range [-19:+19].The camera is set to F# = 2. Illumination is set as office room lighting condition (i.e. 300 lux). Figure 13(c, d) shows some calibrated PSFs in forward and backward point of focus.

In the first experiment, the focal point is set as the farthest point and all objects are placed in front of it. The captured images and results are shown in Figure 14(a). Index number in the color-bar shows relative distance to the camera. Therefore, in each figure, closer object is colored with smaller index.

| (I) Captured image | (II) Depth Map | (III) Deblurred Image |
|---|---|---|

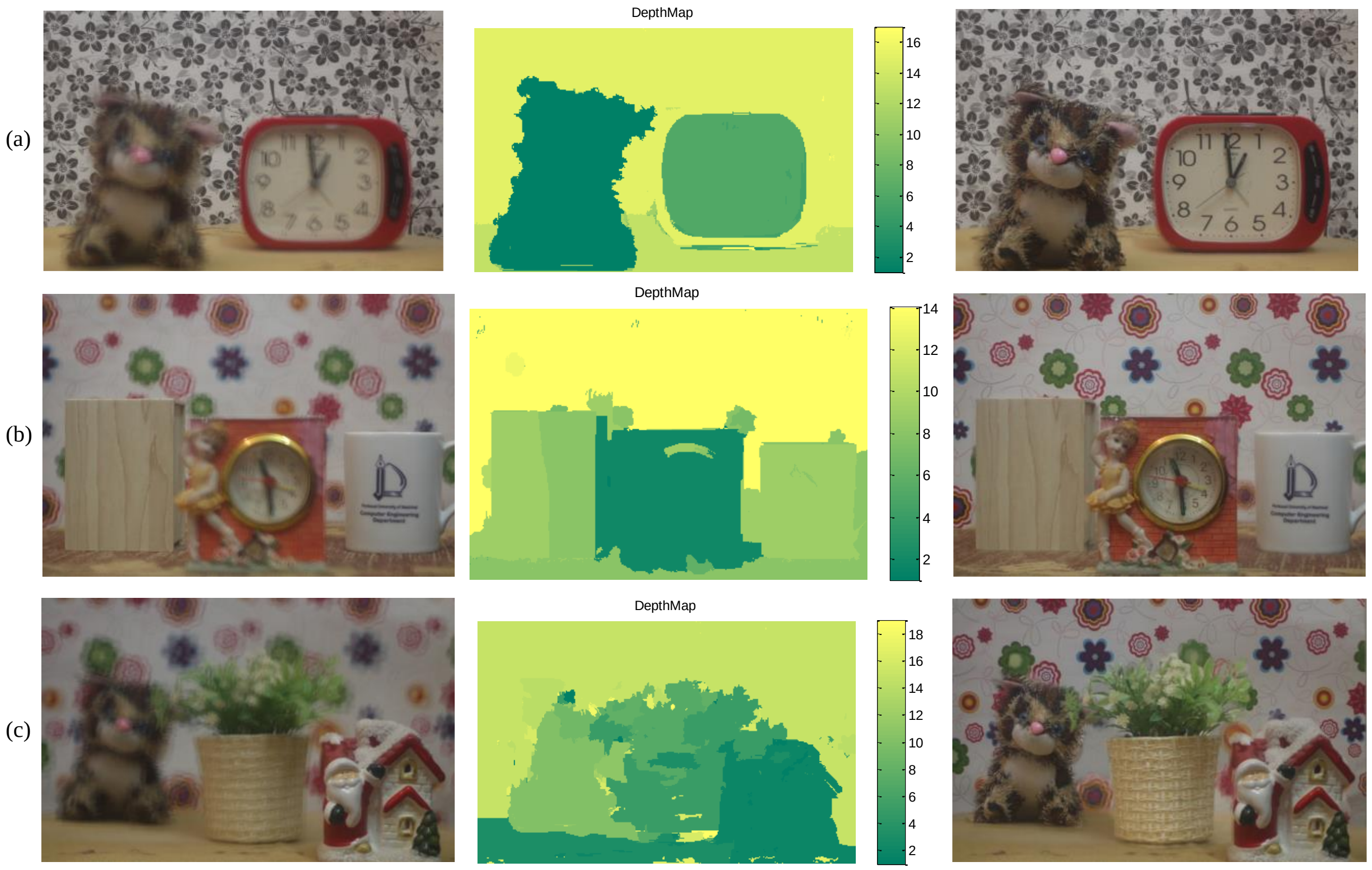

Figure 14Depth map estimation of depth varying scenes: (a) in front of the focal plane, (b) both sides of the focal plane, (d) back of the focal plane.

Although the result is acceptable, there are some errors in depth estimation on the floor of the scene that they should be corrected by user or other segmentation techniques, which are not so sensitive to intensity similarity, should be used.

In the second experiment, three objects are placed in the back of, on and front of the focus point. Figure 14(b) shows the captured image and the obtained depth map. In the third experiment, the focal point is set on the nearest object and all other objects are placed behind that. Figure 14(c) shows our method can also obtain an acceptable result in this case.

Each of the depth-maps are slightly corrected. Then, deblurring[15] is done with the modified depth map. Figure 14(III) shows all-focus images obtained by deblurring.

## 6. Conclusion and future work

In this paper, a new method for aperture mask evaluation was proposed that reduces estimation error in both depth map and deblurring results. Asymmetric apertures make different PSFs in the back and front of the focal point. This feature can help us to discriminate blurred objects which are on two sides of the focal plane. It was also shown that suitable aperture mask might be different in different illumination settings. Our proposed mask was designed for office room illumination setting. To the best of our knowledge, for the first time, aperture evaluation functions were formulated by considering the aperture throughput and imaging conditions. It helps to exact evaluation of masks with different throughput. Analytical and experimental results show that our proposed mask can estimate an appropriate depth map of objects captured in only one image regardless of being in which side of the focal plane. This achievement was obtained with the assistance of a new depth estimation algorithm proposed in this article. According to the proposed algorithm, the deblurring result of correct PSF has the best quality which helps PSF estimation. Although the proposed no-reference quality measure gives good results in depth estimation, more studies could lead to achieving better measures which reduce depth estimation error in both conventional and coded aperture imaging.